\documentclass{article}
\usepackage{spconf,amsmath,graphicx,amssymb}
\usepackage[normalem]{ulem}

\newcommand{\bfx}{\ensuremath{\mathbf{x}}}
\newcommand{\bfy}{\ensuremath{\mathbf{y}}}

\newcommand{\bfM}{\ensuremath{\mathbf{M}}}

\def\min{{\operatorname{min}}}

\title{ROBUST, FAST AND ACCURATE: A 3-STEP METHOD FOR \\ AUTOMATIC HISTOLOGICAL IMAGE REGISTRATION}
\name{Johannes Lotz$^{\star}$, Nick Weiss$^{\star}$\thanks{$^{\star}$JL and NW contributed equally.} and Stefan Heldmann}
\address{Fraunhofer MEVIS \\ 
L\"ubeck, Germany}
\begin{document}
%
\maketitle
\begin{abstract}
We present a 3-step registration pipeline for differently stained histological serial sections that consists of 1) a robust pre-alignment, 2) a parametric registration computed on coarse resolution images, and 3) an accurate nonlinear registration. In all three steps the NGF distance measure is minimized with respect to an increasingly flexible transformation. We apply the method in the ANHIR image registration challenge and evaluate its performance on the training data. The presented method is robust (error reduction in 99.6\% of the cases), fast (runtime 4 seconds) and accurate (median relative target registration error 0.19\%).
\end{abstract}
\section{Introduction}
\label{sec:intro}

In cancer diagnostics and histology related basic research, much insight into molecular and cellular interactions, tissue growth, and tissue organization is gained by analyzing consecutive but differently stained histological sections. For this procedure, a fixed tissue is transferred in a paraffin block and cut into $2-5\mu$m thin slices. Then, slices  are subsequently stained by e.g. immunohistochemistry, and finally examined by a scientist or physician using conventional or virtual microscopy.

In order to correlate the staining intensities, staining patterns, and even subcellular localizations of various proteins or antigens, multiple stainings are frequently required.  To recombine the information from the separate stains, a precise, multi-modal image registration is essential. 

We present a 3-step registration pipeline that consists of 1) a robust pre-alignment, 2) a parametric registration computed on coarse resolution images, and 3) an accurate nonlinear registration.

\section{Related Work}
\label{sec:relatedwork}

The  underlying variational image  registration framework of this work has been described in \cite{fischer_fast_2003,modersitzki_fair_2009} and its application to histological images was first described in \cite{schmitt_image_2006} in 2006. A general issue has been the handling of large images and the associated computational complexity and runtimes. At this time, the elastic registration of two images from slices of a human brain with $512\times 512$ pixels took about 100 minutes on a workstation and 3 minutes on a cluster computer. Later, a faster implementation for regular workstations reducing memory read and write operations has been proposed in \cite{ruhaak_highly_2013} in 2013. The authors report a registration time of 104 seconds for a pair of images from the DIR-Lab 4DCT dataset (approx.  $256\times 256 \times 81$ voxels). Additional optimizations including the exploitation of special instruction sets of modern CPUs has been recently described in \cite{konig_matrix-free_2018-1}, reducing the registration time for two $256\times 256 \times 256$ images to 19 seconds. The present work builds on top of these implementations.

\section{THREE STEPS FOR ROBUST, FAST AND ACCURATE REGISTRATION}
\label{sec:3steps}

Following the framework established by Fischer \& Modersitzki \cite{modersitzki_fair_2009}, 
we formulate image registration as minimization of a suitable objective function 
$J(R, T, y) \rightarrow \min$, where $R : \mathbb R^2 \mapsto \mathbb R$ and 
$T:\mathbb R^2 \mapsto \mathbb R$ are the reference and template images and 
$y:\mathbb R^2 \mapsto \mathbb R^2$ is the wanted transformation function. Central for any suitable objective function is a distance or image similarity measure that quantifies alignment quality. Here we use the Normalized Gradient Fields (NGF) distance measure  \cite{haber_intensity_2007}. In the discrete setting we assume we have 2D images composed from $N$ pixels with uniform size $h$ in each dimension and pixel centers $\bfx_1,...,\bfx_N$. Thus, we use the NGF distance measure given by 
\begin{align*}
\text{NGF}(R,T,y) &:=  \\
 \quad h^2 \cdot \sum_{i=1}^N 1 &- \left( \frac{\langle\nabla T(y(\bfx_i)), \nabla R(\bfx_i) + \varepsilon^2}{\| \nabla T(y(\bfx_i))\|_\varepsilon  \, \| \nabla R(\bfx_i)\|_\varepsilon} \right)^2
\end{align*}
with $\langle \bfx,\bfy\rangle_\varepsilon = \bfx^\top\bfy+\varepsilon^2$, 
$\|\bfx \|_\varepsilon := \sqrt{\langle \bfx,\bfy\rangle_\varepsilon}$,
and the edge parameter $\varepsilon$ steering the sensitivity with respect to strength of edges in images as well as to noise. The NGF distance measure forces the alignment of edges and therefore is based on morphological structures which makes it robust with respect to staining differences \cite{bulten_epithelium_2019}. In general, NGF is suitable for multi-modal image registration.

Minimizing the NGF distance measure is part of all three steps that build up our registration pipeline. To solve the optimization problem we use a multilevel approach, starting with a low image resolution and refining the transformation on higher image resolutions iteratively. This reduces the risk of being tracked in local minima and speeds up the optimization process \cite{haber_multilevel_2006}.

All three registration steps rely on the edge parameter $\varepsilon$, the number of levels $N_{\text{level}}$, the maximum image dimension in x and y for finest level $N_{\text{max}}$ and are set independently for each step.

\subsection{Step 1: Pre-Alignment (Automatic Rotation Alignment)}
\label{ssec:pre-alignment}

After manual tissue processing in the lab, neighboring tissue slices can end up in arbitrary positions on the object plate (such as upside down or turned in various ways). Therefore, we do not make any assumptions on initial tissue positioning and aim to find a rough rigid transformation in a first step, correcting for translation and rotation. The result is then used as an initial guess for a more flexible registration in the second step.

The Automatic Rotation Alignment (ARA) starts by determining the center of mass \cite{beatty_principles_1986} of both images (where each pixel's gray value is used as its mass). The vector pointing from the center of mass of the reference image to the center of mass of the template image is then used as initial translation. Several possible transformation are computed by starting $N_{\text{rot}}$ rigid registrations with different initial rotations, equidistantly sampled from $[0,\, 2\pi]$. From all rigid registrations, the result with the minimal distance measure is selected as intermediate result.

\subsection{Step 2: Parametric Registration}
The second step of the registration pipeline is a parametric registration with an affine deformation model. In 2D, an affine deformation $y$ has 6 degrees of freedom and we set
$$
y(\bfx) = 
\begin{pmatrix} a_{1}&a_{2}\\a_{4}&a_{5}\end{pmatrix}
\bfx
+ 
\begin{pmatrix} a_{3}\\a_{6}\end{pmatrix}
$$
with parameters $a_1,...,a_6\in\mathbb R$. Then, we minimize the objective function
\begin{align*}
J(R, T, y) = \text{NGF}(R,T, y) \rightarrow \min
\end{align*}
with respect to the parameters $a_1,...,a_6$. The resulting transformation is then used as  initial guess for a subsequent non-parametric registration in the last step.
 
We employ an iterative Gauss-Newton optimization and use the parameters from the pre-alignment (translation $t_1, t_2$, rotation angle $\phi$) as initial guess. That is, we setup the initial affine paramters $a_1,\hdots,a_6$ such that
$$
\begin{pmatrix} a_{1}&a_{2}\\a_{4}&a_{5}\end{pmatrix}
= \begin{pmatrix} \cos(\phi) & -\sin(\phi) \\ \sin(\phi) & \cos(\phi) \end{pmatrix} \quad\text{and}\quad
\begin{pmatrix} a_{3}\\a_{6}\end{pmatrix}= \begin{pmatrix} t_{1}\\t_{2}\end{pmatrix}.
$$



\subsection{Step 3: Non-parametric Registration}
   
The final step is a non-parametric image registration. Here, the transformation $y$ is given by
\begin{align*}
y(\bfx) &= \bfx + u(\bfx)  \text{, }  u : \mathbb R^2 \mapsto \mathbb R^2
\end{align*}
with the displacement field $u$. 

Other than in parametric registration, the non-parametric deformation is controlled by an additional term in the objective function, the regularizer. We use curvature regularization \cite{fischer_curvature_2003} with 
$$
\text{CURV}(y) = \frac12(\|\Delta u_1\|_{L_2}^2 + \|\Delta u_2\|_{L_2}^2).
$$
We then minimize the following objective function with respect to the deformation $y$ and displacement $u$, respectively:
\begin{align*}
J(R, T, y) := \text{NGF}(R,T, y) + \alpha \text{CURV}(y) \rightarrow \min,
\end{align*}
where $\alpha>0$ is a regularization parameter, which controls the smoothness of the computed deformation. The regularization parameter  $\alpha$ is manually chosen to provide a smooth deformation  and to avoid topological changes (grid foldings) while being flexible enough to correct local changes improving image similarity.

For our numerical implementation, the displacement field $u$ is discretized with 1st order B-Splines defined on an uniform control point grid with $m$ points. Then we optimize the non-parametric objective function with respect to the control points. To this end, we use a L-BFGS quasi Newton optimization together with multi-level continuation to avoid local minima and to speed up computations.



\subsection{Registration Parameters}

The final set of registration parameters is shown in Table~\ref{tbl:parameters}.

\begin{table}
  \centering
  \caption{Parameters used in the registration pipeline.\label{tbl:parameters}}
  \begin{tabular}{lr}
  \hline
  \textbf{Step 1: Pre-Alignment}               &              \\
  \hline
  Number of rotation angles $N_{\text{rot}}$   &  32          \\
  Maximum image dimension $N_{\text{max}}$     &  200 pixels         \\
  Number of levels $N_{\text{level}}$          &  4           \\
  NGF $\varepsilon$                            &  0.1   \\
  \\
  \hline
  \textbf{Step 2: Parametric Registration}     &              \\
  \hline
  Maximum image dimension $N_{\text{max}}$     & 1000 pixels  \\
  Number of levels $N_{\text{level}}$          & 5            \\
  NGF $\varepsilon$                            & 0.1             \\
  \\
  \hline
  \textbf{Step 3: Non-Parametric Registration} &              \\
  \hline
  Maximum image dimension $N_{\text{max}}$     & 8000 pixels         \\
  Number of levels $N_{\text{level}}$          & 7            \\
  NGF $\varepsilon$                            & 1.0            \\
  regularizer parameter $\alpha$               & 0.1          \\
  number of grid points $m$                    & 257$\times$257           
  \end{tabular}
\end{table}

\section{DATA PREPROCESSING}
Before registration, all images are converted into an in-house multilevel image format based on sqlite\footnote{https://www.sqlite.org}. Without this conversion, the image loading time is increased by about five seconds per registration. 

In addition, all images are converted to gray and inverted while loading.

\section{APPLICATION TO ANHIR CHALLENGE DATA}
\label{sec:typestyle}

\begin{figure}[!b]
\includegraphics[scale=0.315]{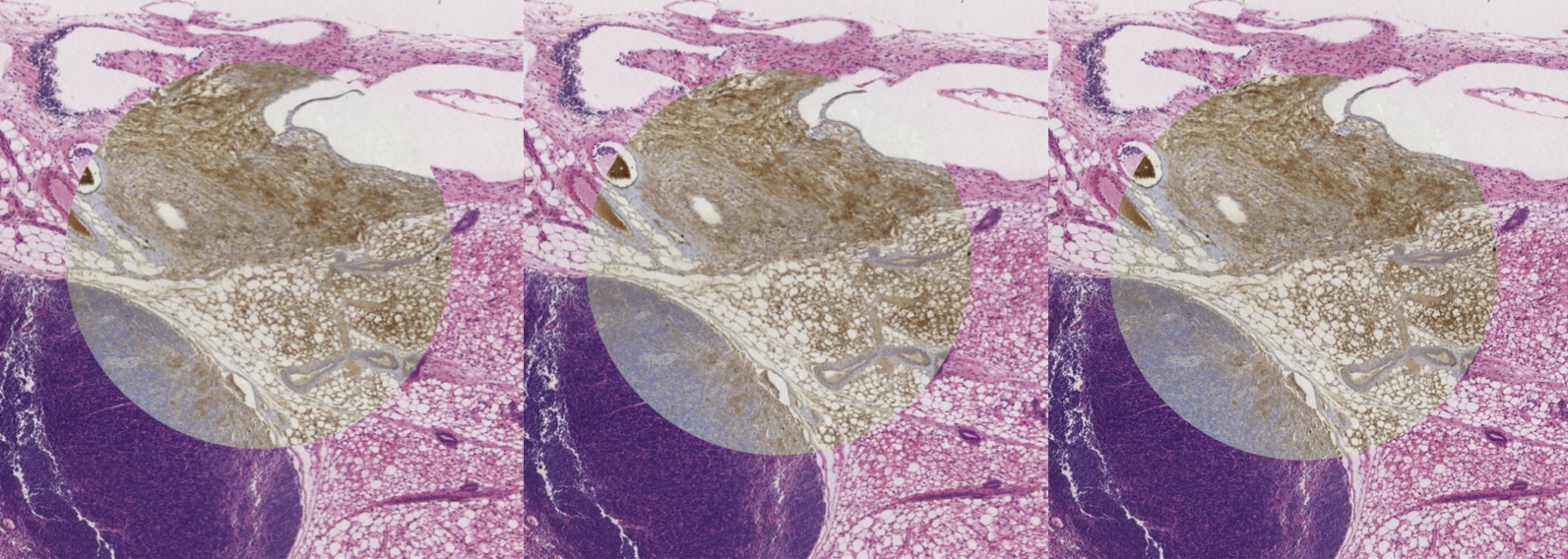}
\caption{Spy-view of an image pair after pre-alignment, parametric and non-parametric registration (left to right).\label{fig:images_anhir}}
\end{figure}

The algorithm has been applied to the data from the ANHIR registration challenge \footnote{https://anhir.grand-challenge.org/Dataset}  \cite{borovec_benchmarking_2018-1,fernandez-gonzalez_system_2002,gupta_stain_2018}. The Registration performance is measured by evaluating the average ($\operatorname{AMrTRE}$) and the median ($\operatorname{MMrTRE}$) of the median relative target registration error ($\operatorname{MrTRE}$) over all image pairs $k=1,...,N_{\text{pairs}}$ following \cite{borovec_benchmarking_2018-1}. The $\operatorname{MrTRE}$ over all landmarks of one image pair $k$ is computed as
\begin{align*}
\operatorname{MrTRE}^k &= \\
\operatorname{median}&\left(\left\{\frac{\|\bfx_l^T - \bfx_l^W\|_2}{\|\bfM\|_2} ,\, l=1,...,N_{\text{landmarks}}\right\}\right)
\end{align*}
where $\bfx^T, bfx^W$ are the template landmarks and the warped reference landmarks and $\bfM \in \mathbb R^2$ is the image extent. $\operatorname{AMrTRE}$ and $\operatorname{MMrTRE}$ are computed as the average and median 
\begin{align*}
\operatorname{AMrTRE} &= \operatorname{mean}(\{ \operatorname{MrTRE}^k ,\, k=1,...,N_{\text{pairs}}\})\\
\operatorname{MMrTRE} &= \operatorname{median}(\{ \operatorname{MrTRE}^k ,\, k=1,...,N_{\text{pairs}}\})
\end{align*}
over the $\operatorname{MrTRE}$s of the image pairs. After registration, landmark errors for the training data $(N_{\text{pairs}}=230)$ of $\operatorname{AMrTRE}=0.49 \%$  and $\operatorname{MMrTRE}=0.19 \%$ are reached. On the subset of pairs where the registration is robust, landmarks errors of $\operatorname{AMrTRE}=0.30 \%$  and $\operatorname{MMrTRE}=0.19 \%$ are reached. 

\begin{figure}
\centering
\includegraphics[width=0.45\textwidth]{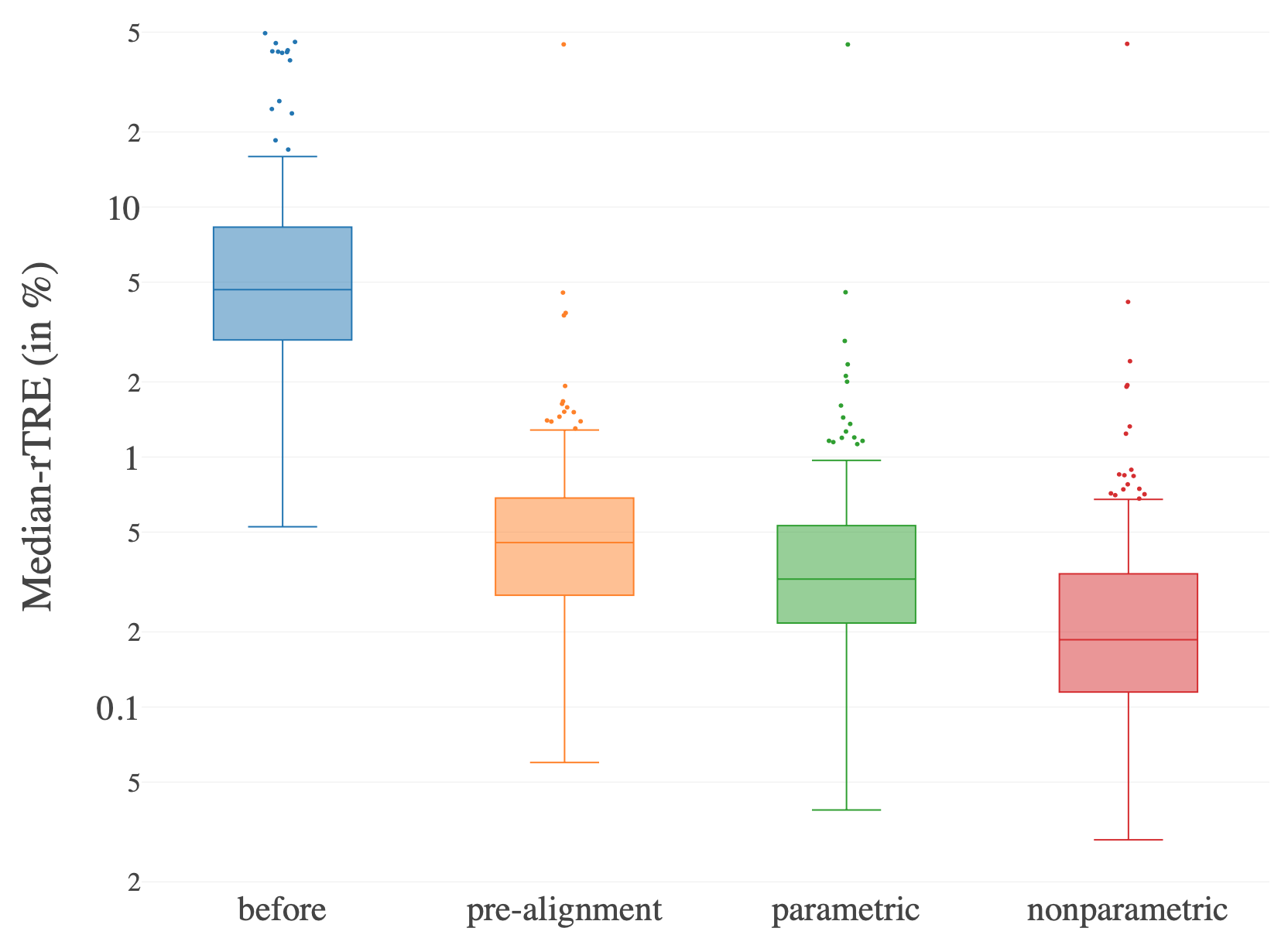}
\caption{Histogram of the Median-rTRE measured on the training data $(N_{\text{pairs}}=230)$ before alignment and after each registration step. \label{fig:histogram}}
\end{figure}

The reduction of the registration error in the training data after each step in the pipeline is shown in the box plots in Figure~\ref{fig:histogram}. While the median error is reduced after each step, those cases that fail in the pre-alignment cannot be recovered at a later stage. Figure~\ref{fig:images_anhir} shows one of the image pairs after pre-alignment, parametric registration and non-parametric registration. 

The resulting deformations do not contain foldings. The average maximum area change in one grid cell was 1.35 \%.

The algorithm is robust in 99.6 \% of the training cases $(N_{\text{pairs}}=230)$. Robustness in the ANHIR challenge is defined as the percentage of the cases where the landmark error is reduced compared to the initial configuration.

Multiple parametrizations were tested on the training data and the parameter set with the lowest median median rTRE (MMrTRE) was selected for submission.

The whole registration process including image loading, pre-alignment, parametric and non-parametric registration takes on average 4.0 seconds on an Intel(R) Core(TM) i7-7700K CPU (4.20GHz, four cores) with 32 GB of RAM.  

\bibliographystyle{IEEEbib}
\bibliography{lotz_weiss_fast_accurate}

\end{document}